\documentclass[letterpaper, 10 pt, conference]{ieeeconf}  
\IEEEoverridecommandlockouts                              

\usepackage{times}

\usepackage{placeins}
\usepackage{multirow}
\usepackage{makecell}
\usepackage{multicol}
\usepackage{hyperref}
\usepackage{diagbox}
\usepackage{url}
\usepackage{lipsum}
\usepackage[utf8]{inputenc} 
\usepackage[T1]{fontenc}    
\usepackage{tgtermes}
\usepackage{wrapfig}
\usepackage{tcolorbox}
\usepackage{booktabs}       
\usepackage{amsfonts}       
\usepackage{nicefrac}       
\usepackage{subcaption}
\usepackage{microtype}      
\usepackage{xcolor}
\hypersetup{
    colorlinks,
    linkcolor={blue!50!black},
    citecolor={blue!50!black},
    urlcolor={blue!80!black}
}

\usepackage{graphicx}

\usepackage{tabularray}
\usepackage{enumerate}
\usepackage{amsmath}

\usepackage{amssymb}
\usepackage{mathtools}

\usepackage{algorithm} 
\usepackage{xspace}
\usepackage{lipsum}

\usepackage{multirow}
\usepackage{makecell}
\usepackage{multicol}
\usepackage{hyperref}
\usepackage{url}
\usepackage{graphicx}
\usepackage{soul}
\usepackage{makecell}
\usepackage{caption}
\usepackage{subcaption}
\usepackage{adjustbox}
\newcommand{\etal}{\textit{et al.}}
\newcommand{\ours}{LANCAR}

\title{\LARGE \bf
LANCAR: Leveraging Language for Context-Aware \\Robot Locomotion in Unstructured Environments
}

\author{Chak Lam Shek$^{1*}$, Xiyang Wu$^{1*}$,  Wesley A.\@ Suttle$^{2}$, Carl Busart$^{2}$, Erin Zaroukian$^{2}$,\\ Dinesh Manocha$^{3}$, Pratap Tokekar$^{3}$, and Amrit Singh Bedi$^{4}$
\thanks{$^*$ Denotes equal contribution}%
\thanks{$^{1}$Department of Electrical and Computer Engineering, University of Maryland, College Park, MD, USA
        {\tt\small \{cshek1, wuxiyang\}@umd.edu}}%
        \thanks{$^{2}$DEVCOM Army Research Laboratory, Adelphi, MD, USA.\@ }%
\thanks{$^{3}$Department of Computer Science, University of Maryland, College Park, MD, USA
        {\tt\small \{dmanocha, tokekar\}@umd.edu}}%
        \thanks{$^{4}$Department of Computer Science, University of Central Florida, Orlando, FL, USA
        {\tt\small \{amritbedi\}@ucf.edu}}%
}

\usepackage{soul}
\begin{document}

\maketitle

\begin{abstract}
Navigating robots through unstructured terrains is challenging, primarily due to the dynamic environmental changes. While humans adeptly navigate such terrains by using context from their observations, creating a similar context-aware navigation system for robots is difficult. The essence of the issue lies in the acquisition and interpretation of context information, a task complicated by the inherent ambiguity of human language.
In this work, we introduce \ours{}, which addresses this issue by combining a context translator with reinforcement learning (RL) agents for context-aware locomotion. \ours{} allows robots to comprehend context information through Large Language Models (LLMs) sourced from human observers and convert this information into actionable context embeddings. These embeddings, combined with the robot's sensor data, provide a complete input for the RL agent's policy network.
We provide an extensive evaluation of \ours{} under different levels of context ambiguity and compare with alternative methods. The experimental results showcase the superior generalizability and adaptability across different terrains.  Notably, \ours{} shows at least a 7.4\% increase in episodic reward over the best alternatives, highlighting its potential to enhance robotic navigation in unstructured environments. 
More details and experiment videos could be found in \href{http://raaslab.org/projects/LLM_Context_Estimation/}{this link}.

\end{abstract}

\section{Introduction}
\label{sec:Introduction}

Designing locomotion for quadruped robots has been a longstanding focus of research~\cite{10.5555/6152}. The variability of environmental physical properties heavily influences the robot's movement, making it difficult to create a universal policy that works in all situations. Despite significant progress, devising a universal policy capable of effectively addressing all possible scenarios remains elusive ~\cite{josef2020deep, xiao2022appl}. 
Previous efforts include using graph-like structures~\cite{liang2023context} or autoencoders~\cite{karnan2023self} to gather context environment information, but these methods often fall short in complex terrain navigation due to limited reasoning capabilities.
%
%
Addressing these challenges requires innovative approaches, such as combining human insights with technological solutions. Humans can intuitively understand environmental contexts, like associating wet grass with high damping, a concept difficult for current algorithms to grasp. 
However, leveraging human feedback is complicated by the ambiguity of natural language~\cite{ahn2018interactive} and the impracticality of expecting humans to provide detailed quantitative descriptions of environments using physical parameters instead of vague qualitative sentences with descriptive words.

\begin{figure*}[t]
    \centering
    \includegraphics[width=0.8\textwidth]{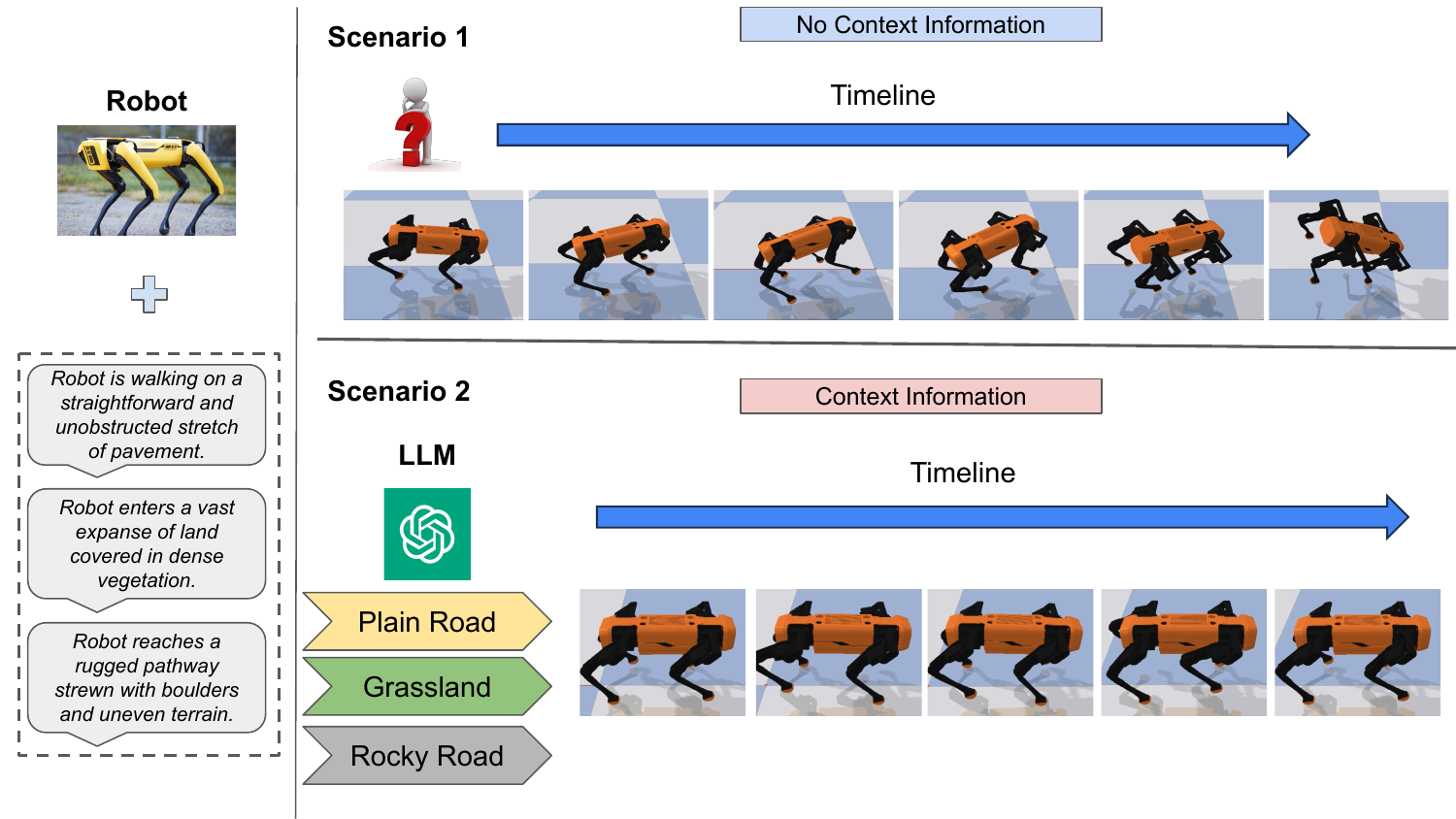}
    \caption{\small{\textbf{Task Description.} We consider two robot learning approaches for locomotion guided by ambiguous human descriptions. The first existing approach (\textbf{TOP}) is when the robot moves over diverse terrains with a trained policy without any context information. Given the complexity of the terrains, robots may face difficulties in developing a generalized policy to address all types of terrains, leading to the failure of its ultimate policy. Our proposed approach (\textbf{BOTTOM}) has the robot moving over diverse terrains with our trained policy and context information from human observers. Robots convert this interpreted context information into embeddings with LLM.\@ With the extra context information added to robots' own perceptions from their sensors, robots could develop better policies with a better understanding of the environment.}}
    \label{fig:task}
    \vspace{-15pt}
\end{figure*}

The recent success of Large Language Models (LLMs) and their ability to perform chain-of-thought~\cite{wei2022chain}, logical reasoning~\cite{creswell2022selection}, and common sense reasoning~\cite{geva2021did} offers a promising solution to these challenges. Techniques such as domain randomization~\cite{tobin2017domain} have been applied to prepare RL agents for varied conditions. Although prior work has explored using LLMs for predicting RL reward functions~\cite{kwon2023reward} or providing robotic control inputs~\cite{mirchandani2023large}, these approaches do not fully utilize LLMs' reasoning capabilities. Positioning LLMs as intermediaries to translate human language into RL-compatible formats could optimize their effectiveness, preventing RL agents' decision-making from being directly influenced by human instructions.

This study explores the use of LLMs to interpret environmental context, enhancing RL agents' ability to guide robot locomotion, particularly for a quadruped robot navigating diverse terrains with the assistance of human observers. The context, often unperceivable directly by the robot, includes terrain characteristics crucial for navigation.
Fig.~\ref{fig:task} gives an overview of our approach. In Scenario $1$, a robot traverses terrains without any context information. The result shows the robot is struggling to formulate a universal navigation policy.
In Scenario $2$, the robot navigates the same terrains but receives additional context information from human observers,\footnote{In some cases vision-language models may also be a useful surrogate to provide context information that is visual} like \textit{``You are entering a grassland right after the rain''} or \textit{``You are walking on a dry rocky road under the sun.''} Robots use an LLM-based translator to convert embeddings representing context information from human interpretation, enhancing their decision-making process alongside their sensory observations.

%
%

Our method, \ours{} (\textbf{LAN}guage for \textbf{C}ontext-\textbf{A}ware \textbf{R}obot locomotion), capitalizes on the versatility of LLMs to interpret human language, transforming it into indices or context embeddings. This process mitigates the ambiguity inherent in human language, enabling robots to navigate varied terrains with adaptable, generalized control policies through collaboration with human observers. While this paper focuses on leveraging LLMs instead of Vision Language Models (VLMs) for context understanding from images, our approach can readily be extended to incorporate VLMs in practical environments.
We validate our approach using the \textit{spot-mini-mini} robot simulator \textit{v.2.1.0}~\cite{spotminimini2020github}, showing that \ours{} enhances performance compared to a no-context baseline. Specifically, \ours{} with context embeddings shows at least 7.3\% improvement on episodic rewards in low-level tasks and 7.5\% improvement in high-level tasks over the baselines. Additionally, \ours{} using the indexing feature leads to a 1.7\% improvement in high-level tasks compared to the no-context approach.


\noindent\textbf{Main Contributions:} We summarize our main contributions in this work as follows.
\begin{itemize}
    \item We propose a novel approach, \ours{}, that incorporates LLMs into RL in robot decision-making that enables robots to understand external context information from human observers and generate a more robust and generalized RL policy.
    \item We propose an LLM-based context information translator module that interprets \emph{high-level}, ambiguous, human language context information of environments into context information embeddings accessible for RL agents with the reasoning ability of LLMs.
    \item We evaluate \ours{} with four different backbone RL approaches under $10$ case studies using both \emph{low-level} and \emph{high-level} context information. We validate the efficacy of \ours{} in policy generalizability and adaptability across diverse terrains which shows at least 7.3\% and 7.5\% of performance improvement over established baselines and over $10$ times higher episodic reward than ablations using different backbone RL approaches.
\end{itemize}

\section{Related Works}
\label{sec:Related_Works}
\noindent \textbf{Robot Navigation in Complex Environments.} 
The challenge of reliable robot locomotion and navigation within complex environments requires adaptive policy learning due to the diversity of terrain encountered~\cite{elnoor2023pronav}. The strategies~\cite{lavalle2006planning,canny1988complexity,manocha1992algebraic} have been developed to address this challenge.
NAUTS~\cite{siva2022nauts} proposes an approach that makes robots adaptive to off-road diverse terrain with a negotiation process among different navigational policies. VINet~\cite{guan2023vinet} uses a novel navigation-based
labeling scheme for terrain classification and generalization
on both known and unknown surfaces. Ada-Nav~\cite{patel2023ada} presents a novel approach that adaptively tunes policy evaluation trajectory lengths with policy entropy and evaluates this approach in both simulated and real-world outdoor environments. Vision-based approaches for terrain adaptation like ViTAL~\cite{fahmi2022vital}, CMS~\cite{loquercio2023learning}, and RMA~\cite{kumar2021rma} use visual observations to generate embeddings that enhance a robot's adaptation capabilities, enabling tasks such as stair climbing or rocky road navigation. However, despite these advances, the current methods are tested within a limited terrain dataset and rely heavily on semantic terrain adaptation strategies, potentially limiting their generalization capabilities in varied real-world terrains.

\noindent \textbf{Human-robot Collaboration.} Human-robot collaboration 
explores how humans and robots can interact effectively to achieve complex tasks, leveraging human cognitive capabilities~\cite{leeper2012strategies}, particularly in unstructured settings~\cite{ajoudani2018progress}.
Such human-in-the-loop collaboration with robotics for tasks like trajectory planning~\cite{dani2020human}, and manipulation~\cite{raessa2020human} in challenging environments like surgery~\cite{fosch2021human} and disaster rescue~\cite{dedonato2015human}. The integration of LLMs has enhanced human-robot collaboration, enabling robots to draw on human knowledge and reasoning. Ren \etal ~\cite{ren2023robots} propose an approach that allows robots to seek help from humans with the assistance of LLM. SayTap~\cite{tang2023saytap} uses foot contact patterns as the interface between human commands in natural language and a locomotion controller that outputs \emph{low-level} commands. RE-Move~\cite{chakraborty2023re} uses human-language instructions to help robots avoid obstacles. LM-Nav~\cite{shah2023lm} uses LLM and VLM in object detection for robots' navigation tasks. These developments underscore the utility of LLMs in robot control, but the challenge of achieving policy generalization across diverse contexts remains unaddressed, marking the primary focus of our investigation.

\noindent\textbf{Language Model for Robotics.} The integration of LLMs~\cite{brown2020language} and Vision Language Models (VLMs)~\cite{zhou2022learning} with robotics marks a significant advancement in embodied AI~\cite{fan2024embodied, dorbala2023can}. This fusion allows robots to leverage the commonsense and in-context learning (ICL) of language models~\cite{dong2022survey} in decision-making tasks~\cite{brohan2023rt, liang2024mtg, padalkar2023open}. Research efforts have enhanced these models' capabilities, such as pre-training for task prioritization \cite{ahn2022can} and converting complex instructions into detailed tasks with rewards \cite{yu2023language}.
RT-2~\cite{brohan2022rt, brohan2023rt} allows manipulators to use the Internet-scale data from the VLMs in their decision-making by taking the action output sequence as another language. Bucker \etal ~\cite{bucker2022reshaping, bucker2023latte} use LLMs to allow human language to improve the manipulator trajectories. Mees \etal ~\cite{mees2023grounding} use LLM to decompose the \emph{high-level} tasks into sub-tasks for the robot to execute. Fu \etal ~\cite{fu2023drive} use LLMs as a driving assistant in autonomous driving tasks. For reinforcement learning, prior works have explored using LLMs in determining reward values~\cite{kwon2023reward} and policy explainability in human-AI interaction~\cite{hu2023language}. Despite these advancements, the specific application of LLMs for interpreting environmental observations and integrating this understanding into RL agents' decision-making processes has not been explored extensively, an area our work aims to address.

\section{Methodology}
\label{sec:Methodology}

\subsection{Problem Formation}

We model the problem as an extension of a Partially Observable Markov Decision Process (POMDP), specifically as an implicit POMDP~\cite{ghosh2021generalization}. An implicit POMDP is specified  by a tuple, $\left \langle\mathcal{S}, \mathcal{A}, \mathcal{O}, \Omega, \mathcal{Z}, \mathcal{F}, \mathcal{T}, \mathcal{R}, \gamma \right \rangle$, where the state space, $\mathcal{S} = \mathcal{S}_{ex} \cup \mathcal{S}_{im}$, is composed of both explicitly observable states $\mathcal{S}_{ex}$ and implicitly observable states $\mathcal{S}_{im}$. The explicitly observable states are those environmental states directly observable from the agent's onboard sensors. The agent's observation space is $\mathcal{O}$. The observation function is given by $\Omega: \mathcal{S}_{ex} \rightarrow \mathcal{O}$.
The implicitly observable states are the context information in the environment that cannot be detected directly by the robots but still affect robots' policies.
$\mathcal{Z}$ denotes the embedding of the context information from the implicitly observable states $\mathcal{S}_{im}$, while the mapping function between the two is $\mathcal{F}: \mathcal{S}_{im} \rightarrow \mathcal{Z}$.
Nevertheless, the implicitly observable states (\textit{i.e.} context information) can still be inferred by robots through reasoning over visual perception or tactile sensing or through human language feedback. In this work, our primary focus is to recover $\mathcal{S}_{im}$ using context information given in natural language. 

The action space $\mathcal{A}$ represents the agent's feasible actions. The transition function $\mathcal{T}: \mathcal{S} \times \mathcal{A} \rightarrow \mathcal{S}$ characterizes the dynamics of the robot within the environment. The reward function $\mathcal{R}: \mathcal{S} \times \mathcal{A} \rightarrow \mathbb{R}$ quantifies the reward of the agent's actions. $\gamma$ is the discounted factor. The agent's policy $\pi$ is given by $\pi: \mathcal{O} \times \mathcal{Z} \rightarrow \Delta(\mathcal{A})$, while $\Delta(\mathcal{A})$ represents the probability distribution over the action space.
We formulate our problem as a finite horizon optimization. The objective is to find an optimal policy $\pi^*$ that maximizes the expected cumulative reward 
\begin{align}
    \pi^* = \arg\max_{\pi} \mathbb{E}_{\pi \sim \{ s_t, a_t \}_{t=0}^{H - 1}} \sum_{t=0}^{H - 1} \gamma^t R(s_t, a_t)
\end{align}
where $H$ is the length of the episode.


\begin{figure}
    \centering
    \includegraphics[width=0.5\textwidth]{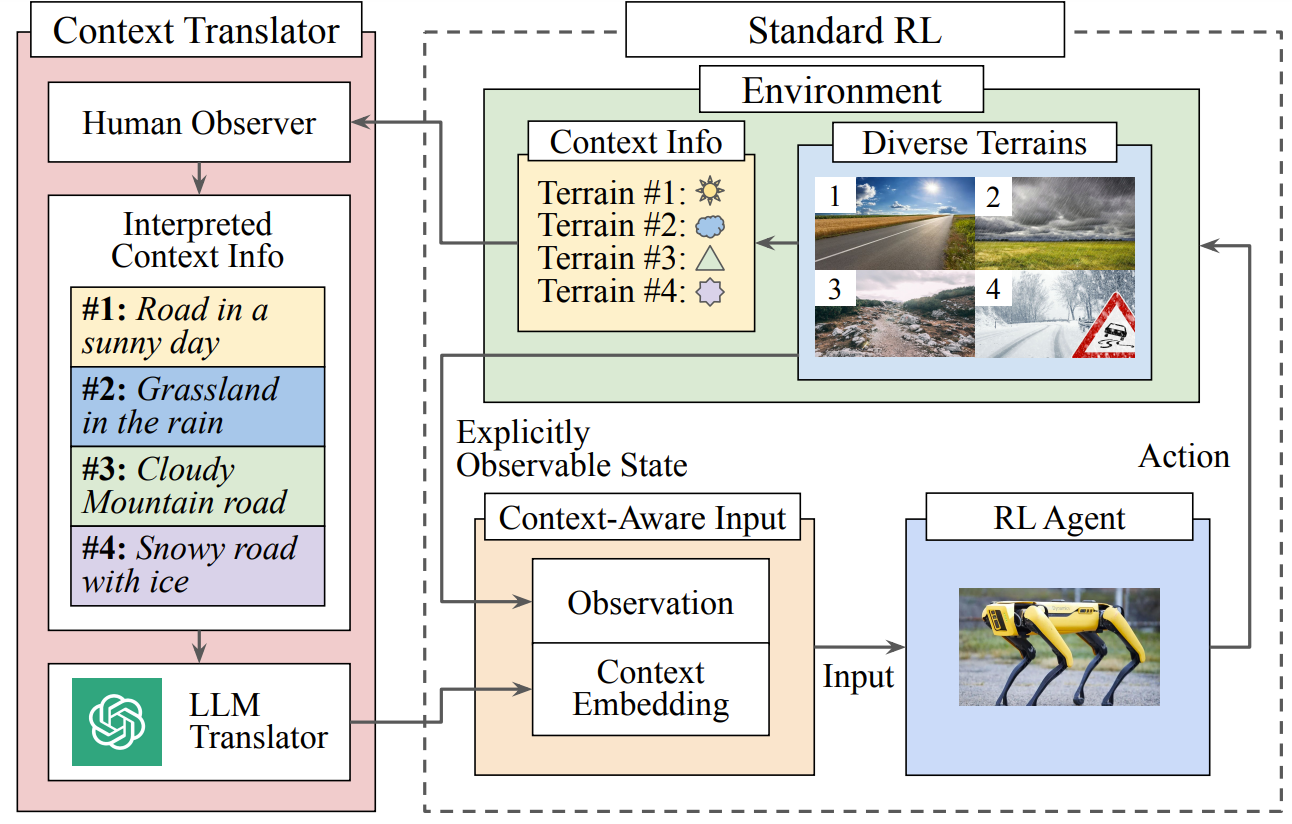}
    \caption{\small{\textbf{Context-Aware Reinforcement Learning Robot Locomotion.} Our framework adds a context translator to the RL setup, enhancing navigation across varied terrains. Agents receive direct observations from the environment, while human observers provide context information, interpreting terrain properties into natural language. The LLM translator processes this into context embeddings, merged with direct observations for RL agent input. The agents then apply their control policies to these enriched inputs to determine and perform actions in the environment.}}
    \label{fig:framework}
    \vspace{-15pt}
\end{figure}

\subsection{Human-Robot Collaboration Framework}
We introduce a human-robot collaboration framework, \ours{}, as depicted in Fig. \ref{fig:framework}.
To recover the context information from environments, we introduce the LLM-based context translator module in addition to the standard RL agent. When robots traverse in environments with diverse terrains at time $t$, robots observe the environment's explicitly observable states $s_{ex}^t$, and the human observer interprets the implicitly observable states $s_{im}$ (\textit{i.e.} context information). Here, we assume that the context information is consistent within one episode so that $s_{im}$ is fixed. The human observer provides qualitative descriptions or captions of the context information to the LLM translator. The LLM translator extracts the environmental properties from the context information and generates the context embedding $z$, which is concatenated with the observations $o_t$ as the input for RL agents. RL agents produce the action $a_t$ using their control policies $\pi$ given the context-aware inputs and execute the action in the environment. This framework is designed to be compatible with the other RL methods, offering flexibility in its implementation.  

The framework is designed delicately to adapt humans' assistance to enhance agents' performance. While it is hypothesized that well-trained agents are better suited to produce a sequence of continuous decisions, direct human control over such well-trained agents may disrupt the decision-making process, potentially leading to degraded performance. On the other hand, human-provided descriptions translated into state estimates over $\mathcal{S}_{im}$ can serve as valuable assistance, enabling the agent to improve its overall performance. 

\subsection{LLM-based Context Translator}
\label{LLM_context}

In our framework, the LLM is pivotal, converting human-interpreted environmental context information into embeddings that RL agents can directly use. We crafted a context translator module leveraging In-context Learning (ICL), enabling the LLM to use its reasoning capabilities through zero-shot or few-shot examples, thus facilitating an interpretable way to interact with the LLM sans explicit training, akin to mimicking human reasoning and decision-making processes~\cite{dong2022survey}.

\begin{figure}
    \centering
    \includegraphics[width=0.48 \textwidth]{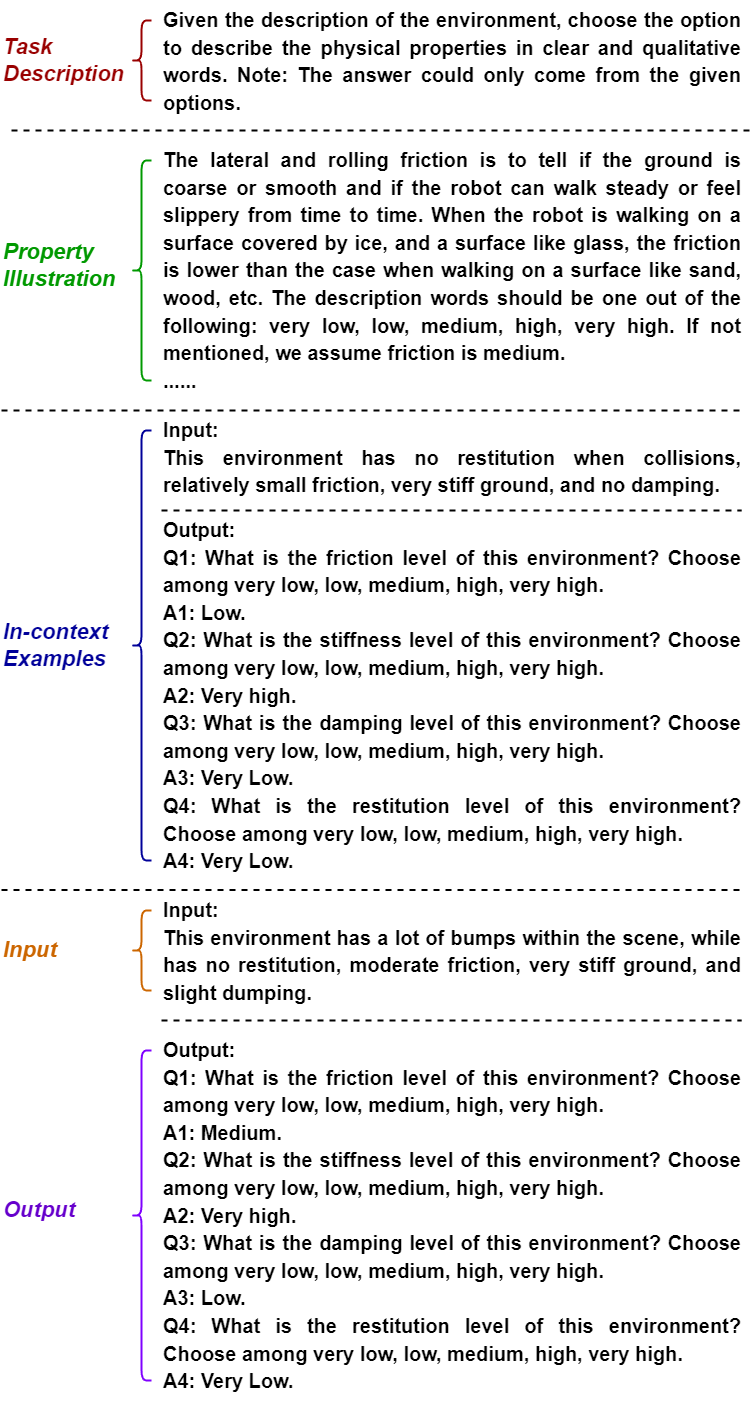}
    \caption{\small{\textbf{An Example Prompt for \ours{}.} The prompt for \ours{} consists of five sections. The first section outlines the \emph{high-level} task for the LLM.\@ The second provides details and examples of relevant terrain properties. The third includes in-context learning examples, featuring \emph{low-level} terrain contexts with outputs derived from multiple-choice question-answering. The final two sections involve presenting inputs to the LLM to generate context embeddings and the corresponding outputs.
    }}
    \label{fig:prompt sample}
    \vspace{-15pt}
\end{figure}

We feed the LLM descriptive sentences about the environment's context information and accompany these with prompts that include examples of potential inputs and outputs the model encounter. An example prompt is presented in Fig. \ref{fig:prompt sample}. These prompts are structured to guide the LLM in mapping qualitative environmental descriptions to embeddings through a series of multiple-choice questions. Each question pertains to a specific environmental characteristic, with the LLM tasked with selecting from pre-defined qualitative descriptors. The chosen answers are then transformed into concatenated one-hot vectors, creating the context embeddings for RL agents.

By providing in-context examples, we aid the LLM in grasping the task's nature. Upon receiving inputs, the LLM is expected to respond to the questions based on the established format of the in-context examples, ensuring the generation of relevant context embeddings.

Let $\oplus: X \times Y \rightarrow [X, Y]$ be an operator concatenating two vectors, and let $\text{onehot}(x)$ denote the function for one-hot encoding. For each property $i$, it could have different levels in a set $P_i$ which include \textit{Very Low}, \textit{Low}, \textit{Medium}, \textit{High} and \textit{Very High}. Given $n$ properties and their corresponding levels $v_{p_i}$ for property $i$, the context embedding of this terrain, denoted as $C$, is obtained by considering the concatenation of the one-hot encoding of the property indexes:
\begin{equation}
    C = \text{onehot}(v_{p_1}) \oplus \text{onehot}(v_{p_2}) \oplus \ldots \oplus \text{onehot}(v_{p_n}) , 
\end{equation}

For example, if the context information describes the terrain with two properties, saying \textit{This terrain has very low friction and very high damping}. \textit{Very low friction} maps into an one-hot vector $[ 1, 0, 0, 0, 0]$ and \textit{very high damping} maps into another one-hot vector $[ 0, 0, 0, 0, 1]$, then the context embedding of this terrain is $[ 1, 0, 0, 0, 0, 0, 0, 0, 0, 1]$.

We note that our context approach, leveraging human-generated prompts and responses, enables the LLM to effectively bridge the gap between natural language descriptions and actionable state information, a key aspect of our framework's success in recovering context information from unobservable states of environments.

\subsection{Reinforcement Learning Agent}
In this work, we employ Augmented Random Search (ARS)~\cite{mania2018simple}, as the reinforcement learning algorithm for the robot control agent. Both ARS and its ancestor approach, BRS, use the finite difference approach, which approximates the gradient value through derivative sampled in $2N$ directions and updates the network parameters by perturbing policy parameters within the range of $[ -\delta, +\delta ]$ to assess resulting rewards within that range, while $\delta$ is randomly generated from a normal distribution. Compared with BRS, ARS further improves the performance of RL policies by normalization and using top-performing directions to update the network parameters. In addition, ARS uses a linear policy, instead of a non-linear policy like the neural network, to simplify the RL algorithms. 

Apart from the ARS approach, we also introduce three widely-used reinforcement learning approaches as our baselines, including SAC~\cite{haarnoja2018soft}, PPO~\cite{schulman2017proximal} and TD3~\cite{fujimoto2018addressing}. More details and discussions are included in Section \ref{sec:result}.

\section{Empirical Results and Discussion}
\label{sec:result}
In the experiments, we aim to answer the following two questions regarding performance and policy generalization. The first question is: \textit{Does external context information improve the performance of the agent when the agent is operating in diverse conditions?} To answer this question, we designed a series of experiments and compared our framework with alternative approaches that include or exclude context information. In these experiments, each context is a different terrain.
The second question we investigate is: \textit{Is the LLM model effective when the given input is \emph{high-level}, open-ended, and ambiguous in retrieving context information and thereby in robot locomotion?}. To answer this question, we use \emph{low-level}, precise, and organized human interpretation of context information in training but also use \emph{high-level}, vague, and unorganized context information from human observers in evaluation, apart from the \emph{low-level} context information evaluation cases. 

\begin{figure*}
    \centering
    \includegraphics[width=\textwidth]{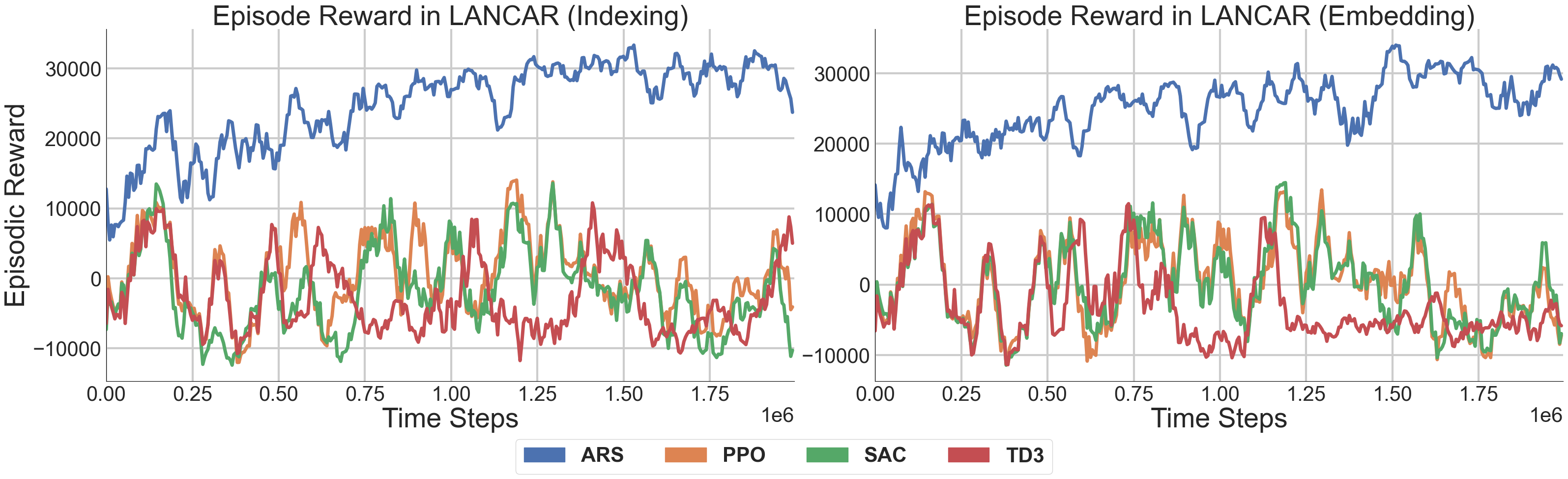}
    \caption{\small{\textbf{Episodic Reward Curve for \ours{} (Indexing) and \ours{} (Embedding) with Different Backbones.} All results are run over $2$ million time steps while each episode has $5000$ time steps in maximum. \textbf{Conclusion.} Both \ours{} (Indexing) and \ours{} (Embedding) have the highest episodic reward when using ARS (blue) backbone than using other backbone approaches.
    }}
    \label{fig:traing_curve}
\end{figure*}

We use GPT-4~\cite{achiam2023gpt} as our LLM model to interpret both  \emph{low-level} and \emph{high-level} human instructions for robots into robot-understandable embeddings with a series of formulated multiple-choice questions. All reinforcement learning agents are trained under domain randomization manner with $8$ different scenarios, all of which have diverse environmental properties. The episodic reward curve during training is present in Fig. \ref{fig:traing_curve}. After the training process, agents are evaluated under $10$ evaluation cases with $5$ cases using \emph{low-level} context information from the environment and $5$ cases using \emph{high-level} context information. All evaluation results are averaged over $16$ episodes.


\subsection{Environments}

We use a quadruped robot locomotion simulator, \textit{spot-mini-mini} \textit{v.2.1.0}~\cite{spotminimini2020github}, built in PyBullet~\cite{coumans2021}. The robot's goal is to advance along the x-axis as much as possible within a fixed set timeframe of 5,000 steps, minimizing deviation from this axis. The raw observation state from the environment is $16$--dimensional, including the robot's roll, pitch, gyroscopes, and acceleration, as well as a $4$-length binary vector denoting the robot's leg contacts with the ground. The extra observation state, depending on the embedding mechanism covered in Section \ref{method name}, is also provided to the agent.
The action space is the desired joint angle for each of the $14$ joints that are clipped within the maximum allowed velocities. The reward function is given by the combination of multiple terms corresponding to the robot's state, including the robot's traveling distance $d_x$, denoting the accomplished distance from the origin in $x$-axis, the penalty $d_y$ that represents the robot's deviation from the $y$ axis, and the penalty $r_p$ that happens when the robot does not keep the desired rate. The reward function is defined as $J = d_x + 0.03d_y + 10r_p$. For the ARS agent using in \ours{}, the learning rate is $0.03$. The number of samples for $\delta$ is $16$. The noise amplitude applied in the exploration is $0.05$.


\begin{table}[h]
    \caption{\small{Properties for Training Terrains}}
    \label{table_phy_properties}
    \begin{center}
    \begin{tabular}{|c||c|}
    \hline
    \textbf{Property} &\textbf{ Value Range} \\
    \hline
    Restitution & $[0, 0.2]$\\
    \hline
    Lateral / Horizontal Friction & $[0, 1]$\\
    \hline
    Rolling Friction & $[2 \times 10^4, 1.6 \times 10^5]$\\
    \hline
    Stiffness & $[0, 1]$\\
    \hline
    Damping Coefficient & $[0, 0.5]$\\
    \hline
    \end{tabular}
    \end{center}
    \vspace{-12pt}
\end{table}

Manually crafting context information for each training instance is unfeasible due to the vast amount of data. Instead, we employ an LLM to automate the generation of detailed, low-level terrain descriptions. This process starts with generating random samples of the terrain's parameters quantitatively describing properties given in Table~\ref{table_phy_properties}, followed by a prompt instructing the LLM to translate these values into qualitative descriptions—ranging from \textit{Very Low} to \textit{Very High}, based on $18$ in-context learning examples. Given the actual parameter values, we ask the LLM to generate a \emph{low-level} context description. A sample description generated during training is: \textit{This environment has very low restitution when collision, low friction, very high stiffness level, and very low damping.}

\subsection{Evaluation Cases}
In the evaluation phase, we conduct two case study experiments with increasing difficulty levels, \textit{i.e.} increasing vagueness of the context information provided to the LLM, to examine the reasoning ability of our approach. Specifically, we evaluate the following two types of contexts (in increasing order of open-endedness):
\subsubsection{Low-Level Context} The context information provided by human observers during evaluation gives detailed qualitative descriptions of environmental properties, the same as those given in the training phase. Descriptions for all five evaluation cases we used are provided in Table~\ref{low-level scenes}.
\begin{table}[h]
    \caption{\small{Low-Level Context Information for Case Study}}
    \label{low-level scenes}
    \begin{center}
    \begin{tabular}{|c||c|}
    \hline
    \textbf{ID} & \textbf{Context Information}\\
    \hline
    A  & \makecell{"This environment has no restitution when \\ collision, very high friction, and no damping."}\\
    \hline
    B &  \makecell{"This environment has no restitution when \\ collision, very low friction, and no damping."}\\
    \hline
    C & \makecell{"This environment has high restitution when \\ collision, very high friction, and very high damping."}\\
    \hline
    D & \makecell{"This environment has medium restitution when \\ collision, low friction, and very high damping."} \\
    \hline
    E & \makecell{"This environment has high restitution when \\ collision, very high friction, and low stiffness."} \\
    \hline
    \end{tabular}
    \end{center}
    \vspace{-12pt}
\end{table}

\subsubsection{High-Level Context} The context description provided by human observers is \emph{high-level}, open-ended, vague, and descriptive of the environmental conditions, as opposed to the environmental properties. Descriptions for all five evaluation cases we used are provided in Table~\ref{high-level scenes}.

\begin{table}[h]
    \caption{\small{High-Level Context Information for Case Study}}
    \label{high-level scenes}
    \begin{center}
    \begin{tabular}{|c||c|c|}
    \hline
    \textbf{ID} & \textbf{Name} & \textbf{Context Information}\\
    \hline
    F & \makecell{Moist \\ Grassland} & \makecell{"The spot is walking on a grassland \\ under a drizzle."}\\
    \hline
    G & \makecell{Snowy \\ Mountain Road} & \makecell{"The spot is walking on a mountain road \\ covered by ice. It's snowy now."}\\
    \hline
    H & \makecell{Sunny \\ Beach} & \makecell{"The spot is walking on the beach \\ near the sea under the sun."}\\
    \hline
    I & \makecell{Rainy \\ Concrete Road} & \makecell{"The spot is walking on a concrete road \\ under heavy rain."} \\
    \hline
    J & \makecell{Sunny \\ Running Tracks} & \makecell{"The spot is walking on running tracks \\ on a sunny day."} \\
    \hline
    \end{tabular}
    \end{center}
    \vspace{-12pt}
\end{table}


\begin{table*}[h]
   
  \caption{\textbf{Average Episodic Reward of \ours{} and No-context RL.} We perform evaluation experiments across all baselines and ablation studies over $10$ cases ($5$ \textit{low-level} context cases and $5$ \textit{high-level} context cases). \textbf{Conclusion.} ARS-based approaches achieve much higher episodic rewards than all other baselines. ARS using \ours{} embeddings for context information have a better performance than all other approaches in most cases.}
  \begin{center}
  \resizebox{0.99\textwidth}{!}{
    \begin{tabular}{lccccccccccc}
    \toprule
        \multicolumn{2}{c}{} & \multicolumn{5}{c}{\makecell{Low-Level Context}}
        & \multicolumn{5}{c}{\makecell{High-Level Context}}  \\
        \cmidrule(lr){3-7} \cmidrule(lr){8-12}
        Method & Backbone  & A & B & C & D & E & F & G & H & I & J \\
        \midrule
        \multirow{4}*{\makecell{{No-Context} \\ ($\times 10^3$)}} & \makecell{ARS} & 36.628 & 19.698 & 38.000 & 28.573 & 30.744 & 35.545 & 13.051 & 29.819 & 34.053 & 33.934 \\
        ~ & \makecell{SAC} & 24.189 & -10.128 & 15.571 & -10.839 & -11.457 & 9.461 & -7.123 & -10.076 & 18.252 & -3.994 \\
        ~ & \makecell{TD3} & 25.001 & -6.756 & 17.768 & -12.230 & -11.726 & 9.833 & -9.445 & -12.450 & 19.352 & -3.583 \\
        ~ & \makecell{PPO} & 7.542 & -8.266 & -1.249 & -10.159 & -10.073 & 4.534 & -7.262 & -10.637 & 15.798 & -2.181 \\
         \midrule
        \multirow{4}*{\makecell{\textbf{\ours{} (Indexing)} \\ ($\times 10^3$)}} & \makecell{ARS} & 36.659 & \textbf{23.435} & 38.366 & 20.649 & 22.952 & 37.791 & \textbf{16.265} & 22.776 & 36.676 & 35.357 \\
        ~ & \makecell{SAC} & 16.423 & -9.695 & 14.534 & -12.199 & -12.443 & 7.521 & -7.592 & -12.012 & 16.252 & -5.815 \\
        ~ & \makecell{TD3} & 20.867 & -7.665 & 15.734 & -11.672 & -11.612 & 7.955 & -7.328 & -13.414 & 17.089 & -4.131 \\
        ~ & \makecell{PPO} & 24.119 & -8.343 & 11.851 & -8.520 & -9.498 & 10.937 & -10.980 & -10.333 & 19.934 & -2.009 \\
         \midrule
        \multirow{4}*{\makecell{\textbf{\ours{} (Embedding)} \\ ($\times 10^3$)}} & \makecell{ARS} & \textbf{41.220} & 20.706 & \textbf{41.725} & \textbf{29.545} & \textbf{31.595} & \textbf{40.563} & 12.162 & \textbf{30.961} & \textbf{39.722} & \textbf{36.623}\\
        ~ & \makecell{SAC} & 12.154 & -8.648 & 17.251 & -9.413 & -11.159 & 8.381 & -7.197 & -12.599 & 16.176 & -5.970 \\
        ~ & \makecell{TD3} & 20.714 & -8.655 & 17.788 & -9.138 & -11.022 & 8.587 & -6.465 & -12.478 & 15.772 & -16.800 \\
        ~ & \makecell{PPO} & 12.979 & -9.449 & 5.512 & -9.187 & -10.314 & 8.345 & -9.533 & -9.391 & 15.607 & -8.148 \\
    \hline
    \end{tabular}
    }
    \vspace{-15pt}
    \end{center}
     \label{lancar_results}
\end{table*}

\subsection{Baselines}
\label{baselines}
We have considered the following baseline algorithms: Augmented Random Search (ARS)~\cite{mania2018simple}, Soft Actor-Critic (SAC)~\cite{haarnoja2018soft}, Proximal Policy Optimization (PPO)~\cite{schulman2017proximal}, and Twin Delayed DDPG (TD3)~\cite{fujimoto2018addressing}. These algorithms are commonly applied to solve a variety of control tasks. ARS is a derivative-free optimization algorithm that explores the parameter space through random perturbations to improve policy performance. SAC leverages an off-policy approach to optimize the policy while also estimating the value function. PPO employs a policy gradient method with a clipping mechanism to ensure smooth policy updates and prevent large policy changes. TD3 is an extension of the Deep Deterministic Policy Gradient (DDPG) algorithm~\cite{lillicrap2015continuous}, integrating twin critics and target policy to address overestimation bias and improve the robustness of the learned policy.

\subsection{Ablation Study}
\label{method name}
We conduct a series of experiments on our approach, \ours{}, and some baseline approaches, to evaluate the effect of the usage and design of context information embedding strategies. We evaluate the following approaches:

\subsubsection{No-Context} The RL agent only uses environmental observation in their decision-making. No context information is used by the RL agent. The decision does not rely on the LLM output. It will be used as the baseline of the experiment.  

\subsubsection{\ours{} (Indexing)} The context is encoded as a one-hot vector identifying the environment. The RL agent labels all terrains encountered during training with a unique index. The one-hot vector sets the $i$-th element of the vector as $1$ and is used as the embedding for the RL agent, denoting the $i$-th training terrain.
During the evaluation, the indexing embedding vector is replaced with an all-zero vector as padding for the RL agent.

\subsubsection{\ours{} (Embedding)}  This is the approach we propose in this work. The LLM generates context embeddings by interpreting human language instruction in the way presented in Section \ref{LLM_context}, and the RL agent incorporates context embeddings with environmental observation in their decision-making. The context embeddings are represented with a combination of multiple one-hot vectors. Each one-hot vector quantifies properties in Table \ref{table_phy_properties} into five intervals.

\subsection{Results}
\subsubsection{Case Study: Low-Level Context Information}

The evaluation involved testing with low-level context information across five different terrain cases: normal terrain (Case A), low friction (Case B), high damping (Case C), medium restitution with very high damping (Case D), and high restitution and damping with low stiffness (Case E). Generally, terrains with low friction, low stiffness, and high damping present greater challenges for RL-controlled robots, with restitution level variations increasing locomotion task uncertainty.

Table~\ref{lancar_results} (Case A-E) shows the evaluation results of episodic reward over all cases using \emph{low-level} context information. We find that \ours{} (Embedding) using an ARS backbone and Embedding method outperformed the other two approaches. \ours{} (Embedding) achieved 16.0\% higher episodic rewards than \ours{} (Indexing) and 7.3\% higher episodic rewards than the No-Context baseline. The \ours{} (Indexing) approach, while slightly underperforming compared to the No-Context baseline by 7.5\% in episodic reward, showed variable performance across terrains, outperforming \ours{} (Embedding) in the low friction scenario (Case B) but falling behind in all other cases. This variability suggests a limited adaptation capability of the Indexing method to different terrains and context inputs. Notably, all methods employing an ARS backbone demonstrated better performance (approximately $17.0\times$ higher episodic reward) than those with other backbones across all evaluation scenarios, underscoring the superior adaptability of approaches with ARS backbone in this context.



\subsubsection{Case Study: High-Level Context Description}
In our second case study, we explored \emph{high-level} context scenarios: Moist Grassland (Case F), Snowy Mountain Road (Case G), Sunny Beach (Case H), Rainy Concrete Road (Case I), and Sunny Running Tracks (Case J). These terrains, characterized by complex combinations of surface properties like stiffness, damping, and friction, present more challenging conditions than those in the \emph{low-level} context study.



Table~\ref{lancar_results} (Case F-J) shows the evaluation results of episodic reward over $5$ \emph{high-level} evaluation cases.
Evaluation results show that \ours{} (Embedding) with an ARS backbone performed better than both \ours{} (Indexing) with 9.3\% higher episodic rewards and the No-Context setups with 7.5\% higher episodic rewards. \ours{} (Indexing) surpassed \ours{} (Embedding) in Case G (Snowy Mountain Road), indicating particular adaptability to low-friction conditions, but \ours{} (Embedding) excelled in all other scenarios. Approaches using the ARS backbone consistently outperformed those with different backbones across all tested terrains. Furthermore, the heightened challenge of these high-level context terrains accentuated the performance gap between ARS-backed methods and other approaches.

\subsection{Discussion}
Our experiments showed that context-aware strategies, \ours{} (Indexing) and \ours{} (Embedding), consistently surpassed the baseline no-context approach in performance. \ours{} (Embedding) generally yielded superior results across diverse scenarios, showcasing its adaptability, whereas \ours{} (Indexing) excelled in specific situations. This discrepancy is attributed to the limited range of training scenarios, which may not encompass a broad spectrum of physical properties, leading to some evaluation cases falling outside the training domain. Expanding the variety of training terrains could address this issue by enhancing the model's exposure to different environments. Notably, \ours{} (Embedding) demonstrated a more significant performance boost over no-context baselines with high-level contexts, suggesting that the LLM's interpretation of context information effectively mitigates environmental ambiguity and the complexities of locomotion tasks. Additionally, methods utilizing the ARS backbone outperformed all alternatives across every scenario, affirming their superior adaptability across a wide array of evaluation conditions.



\section{Conclusion}
This paper introduces a method allowing human observers to use natural language for conveying environmental context to robots, with LLMs translating this into context embeddings for RL agents. These embeddings, combined with the agents' observations, enhance navigation strategies.

Looking ahead, we aim to evolve our methodology by incorporating visual sensors and foundation models for interpreting environmental context, aiming for more accurate object captions within robot-perceived images. Another potential extension of our work is to leverage multi-modal foundation models for context understanding through different sensors, directly creating embeddings understandable by robots. Besides, we plan to explore mechanisms for enhancing robot adaptability across different contexts within the same episode, particularly for outdoor navigation tasks, aiming to improve the robustness and adaptiveness of real-world robot locomotion strategies.

\footnotesize{
\bibliographystyle{ieeetr}
\bibliography{iclr2024_conference}
}


\end{document}